\documentclass[12pt,a4paper]{article}

\usepackage[utf8]{inputenc}
\usepackage[T1]{fontenc}
\usepackage{times}
\usepackage[margin=1in]{geometry}
\usepackage{graphicx}
\usepackage{booktabs}
\usepackage{multirow}
\usepackage{tabularx}
\usepackage{amsmath}
\usepackage{amssymb}
\usepackage{hyperref}
\usepackage{natbib}
\usepackage{xcolor}
\usepackage{enumitem}
\usepackage{float}
\usepackage{tikz}
\usetikzlibrary{arrows.meta, positioning, shapes.geometric}
\usetikzlibrary{calc, arrows.meta}
\usepackage{setspace}
\title{Agent-Driven Corpus Linguistics: A Framework for Autonomous Linguistic Discovery}

\author{
  Jia Yu\textsuperscript{1}, Weiwei Yu\textsuperscript{1}, Pengfei Xiao\textsuperscript{2}, Fukun Xing\textsuperscript{1,*}\\
  \\
  \textsuperscript{1}Zhejiang International Studies University, Hangzhou, China\\
  \textsuperscript{2}Tianjin University, Tianjin, China\\
  \\
  \textsuperscript{*}Corresponding author: \texttt{xingfukun@zisu.edu.cn}
}

\date{}

\begin{document}

\maketitle

\begin{abstract}

  Corpus linguistics has traditionally relied on human researchers to formulate hypotheses, construct queries, and interpret results---a process that demands both
   specialized technical skills and considerable time. We propose \textbf{Agent-Driven Corpus Linguistics}, an approach in which a large language model (LLM),    
  connected to a corpus query engine through a structured tool-use interface, takes over the investigative cycle: it generates hypotheses, queries the corpus,
  interprets the returned data, and refines its analysis across multiple rounds. The human researcher sets the research direction and evaluates the final output. 
  Unlike unconstrained LLM generation, every finding the agent produces is anchored in verifiable corpus evidence. We treat this not as a replacement for the   
  corpus-based/corpus-driven distinction but as a complementary dimension: it concerns \textit{who} conducts the inquiry, not the epistemological relationship
  between theory and data.

  We demonstrate the framework with a concrete implementation that links an LLM agent to a CQP-indexed Gutenberg corpus (5 million tokens) via the Model Context  
  Protocol (MCP). Given only the directive ``investigate English intensifiers,'' the agent identified and quantified a diachronic relay chain (\textit{so}+ADJ
  $\to$ \textit{very} $\to$ \textit{really}), three pathways of semantic change (delexicalization, polarity fixation, and metaphorical constraint), and           
  register-sensitive distribution patterns. A controlled baseline experiment---the same LLM with and without corpus access---shows that grounding contributes   
  quantification, falsifiability, and data-responsive synthesis that the model cannot produce from training data alone. To test external validity, the agent
  replicated two published studies on the CLMET corpus (40 million tokens): \citet{claridge2025reader}'s diachronic decline of \textit{reader} and
  \citet{desmet2013spreading}'s gerund complement spreading pattern, in both cases with close quantitative agreement. These results show that agent-driven corpus
  research can produce empirically grounded, externally validated findings at machine speed---lowering the technical barrier and making corpus methods accessible
  to a broader range of researchers.

\end{abstract}

\section{Introduction}                                                                                                                                          
           
Corpus linguistics rests on a straightforward bet: let large collections of naturally occurring text, rather than linguist intuitions, arbitrate theoretical claims.
  Three decades of tool-building have largely made good on that bet---concordancers, taggers, and parsed corpora now surface regularities in structure, variation, and
  change that introspection alone misses \citep{sinclair1991corpus, tognini2001corpus}.

The bottleneck, though, has barely moved. Whether the work is corpus-based or
  corpus-driven, a human still anchors every stage of the pipeline: choose what to query, scan the output, judge which patterns matter, circle back. CQP and its
  relatives are far from trivial to learn; even a practised analyst can only test so many hypotheses before the hours run out.This creates two practical constraints: a   
  \textit{technical barrier}, since corpus query languages such as CQP require specialized training; and a \textit{time-and-attention constraint}, since the    
  number of hypotheses and cross-tabulations that a single researcher can pursue within any one study is inevitably bounded by the hours available.

  Large language models (LLMs) with tool-use capabilities now make it possible to address both constraints. Current models can generate corpus queries from
  natural-language descriptions, interpret structured output, and---crucially---formulate new hypotheses in response to observed data. When connected to a corpus
  engine through a standardized protocol, an LLM is no longer just a faster query assistant; it becomes an autonomous research agent that explores the corpus on
  its own initiative, subject to human oversight of the research agenda and final evaluation of the results.

  In this paper, we propose \textbf{Agent-Driven Corpus Linguistics}, a methodological approach in which the cognitive work of corpus investigation is delegated
  to an AI agent. Building on \citet{tognini2001corpus}'s distinction between corpus-based and corpus-driven research, we introduce a complementary dimension:
  rather than changing the epistemological orientation, we change \textit{who} performs the inquiry---substituting a human analyst with an AI agent that can
  operate in either mode. We make four contributions:

  \begin{enumerate}
  \item We define the agent-driven approach and position it as a complementary dimension to existing corpus linguistics methodology---one that concerns who
  conducts research, not how theory relates to data. The approach rests on three properties: grounding in corpus evidence, iterative autonomy, and expanded       
  analytical coverage.
                                                                                                                  
  \item We present a working system implementation connecting an LLM to a CQP corpus engine via the Model Context Protocol (MCP), demonstrating that the framework
   is realizable with current technology. The framework itself is protocol-agnostic.
                                                                                                                
  \item We report a case study on English intensifiers in which the agent autonomously produced findings on diachronic change, semantic delexicalization, and     
  register sensitivity. To test generalizability, we replicated two published studies on an independent corpus (CLMET), with closely matching results.
                                                                          
  \item We conduct a controlled baseline experiment comparing the same LLM with and without corpus access, empirically distinguishing what grounding adds beyond  
  training-data recall.
  \end{enumerate}                                                                                                                                                 
                                                                                                        
  The remainder of this paper is organized as follows. Section~\ref{sec:background} reviews related work in corpus linguistics methodology and AI-assisted        
  research. Section~\ref{sec:framework} presents the agent-driven framework, including its formal definition, system architecture, and workflow.
  Section~\ref{sec:case_study} reports the case study, baseline experiment, and replication of published research. Section~\ref{sec:discussion} discusses the    
  implications and limitations of the approach, and Section~\ref{sec:conclusion} concludes.
\section{Background and Related Work}
\label{sec:background}

\subsection{Paradigms in Corpus Linguistics}

The distinction \citet{tognini2001corpus} drew between ``corpus-based'' and ``corpus-driven'' research still supplies the field's main methodological axis.
  Corpus-based work starts from a theoretical framework and uses corpus evidence to test it; the corpus-driven route sets prior categories aside and lets patterning
  emerge from the data. Most practitioners grant the corpus-driven approach a firmer empirical footing, though at a well-known cost in labour                           
  \citep{sinclair1991corpus}.
                                                                                        
  Neither route, however, escapes a practical ceiling. A practised scholar might run a dozen hypothesis tests in one case study; a richly tagged corpus quietly makes   
  hundreds available. The mismatch is not one of competence---we still depend on the researcher's theoretical eye and critical judgement---but of sheer bandwidth: the
  ``observation--interpretation--correction'' loop can only turn so fast when a single person drives it.

\subsection{AI-Assisted Corpus Research}

Several recent systems gesture toward automating empirical research, but none yet covers the full pipeline for corpus linguistics. \textbf{The AI Scientist}
  \citep{lu2024aiscientist} comes closest in ambition: it chains hypothesis generation, experiment execution, and paper drafting into a single loop---yet it targets    
  machine-learning benchmarks exclusively. Corpus work demands structured query languages, layered annotation metadata, and interpretive judgements that have no
  counterpart in a standard ML experiment; AutoRA (behavioural science) and AutoResearcher (literature synthesis) face the same domain mismatch. The one agent that does
   operate on corpus data, \textbf{UDagent} \citep{klemen2025towards}, takes a far narrower brief. A user poses a linguistic question; the LLM translates it into Python,
  runs the code against a Universal Dependencies treebank, and returns the numbers. That the LLM can mediate corpus access at all is noteworthy, but the workflow stays
  strictly linear---no self-generated hypotheses, no iterative follow-up.

  A separate line of work embeds generative models inside existing corpus platforms rather than building standalone agents. \citet{davies2025corpora} showed that GPT
   and Gemini can surface more insightful collocates than legacy frequency lists, despite having no direct access to keyness or collocation statistics.
  \citet{anthony2025antconc} responded by wiring a ChatAI module into AntConc for retrieval-augmented queries over user corpora; \citet{cheung2025corpuschat} built     
  CorpusChat toward a similar end, focused on academic writing support. These integrations are useful, but \citet{uchida2024using} found that current LLMs, while adept
  at spotting broad trends, still cannot match traditional tools on rigorous quantitative tasks. In every case the AI serves as a powerful assistant---it has not yet
  been asked to set the research agenda itself.

  Table~\ref{tab:systems} situates these systems along a spectrum of autonomy, comparing their capabilities across five
  dimensions central to the agent-driven research cycle.

\begin{table}[ht]
\centering
\caption{Comparison of AI-assisted corpus research systems along the autonomy spectrum.}
\label{tab:systems}
\small
\begin{tabularx}{\textwidth}{lXcccc}
\toprule
\textbf{System} & \textbf{Corpus access} & \textbf{Hyp.\ gen.} & \textbf{Query} & \textbf{Interpret} & \textbf{Iterate} \\
\midrule
UDagent & UD treebanks (Python) & --- & \checkmark & --- & --- \\
AntConc+ChatAI & User-uploaded & --- & \checkmark & partial & --- \\
CorpusChat & User-uploaded & --- & \checkmark & partial & --- \\
AI Scientist & ML benchmarks (code) & \checkmark & \checkmark & \checkmark & \checkmark \\
\textbf{Ours} & \textbf{CQP corpora (tool-use)} & \checkmark & \checkmark & \checkmark & \checkmark \\
\bottomrule
\end{tabularx}
\\[4pt]
\footnotesize Hyp.\ gen.\ = autonomous hypothesis generation; Iterate = autonomous multi-round refinement. AI Scientist targets ML, not linguistics.
\end{table}

This landscape can be viewed as a divide. On the ``intelligent assistant'' side, tools like \textbf{UDagent} and \textbf{AntConc+ChatAI} essentially act as technical bridges; they handle the heavy lifting of query translation, yet the human researcher still maintains full agency over the research agenda. The leap toward true \textit{autonomous inquiry} is a different matter. In such a setup, the AI does not merely execute commands---it self-directs the entire cycle, from initial hypothesis to final interpretation. While \textit{The AI Scientist} \citep{lu2024aiscientist} has already proven this level of autonomy is possible for ML benchmarks, its methodology has not yet been adapted for the metadata-heavy, nuanced world of linguistic corpora. Our proposed framework addresses this gap by occupying this latter position: the agent autonomously generates hypotheses, constructs corpus queries, interprets results, and iterates toward findings.

The gap we address is the absence of such an autonomous inquiry system within the methodological context of corpus linguistics.

\subsection{Intelligent Agents and LLM Tool Use}

Intelligent agents---systems that perceive, plan, and act autonomously---date back to early AI research \citep{wooldridge1995intelligent, russell2020artificial}. What makes LLM-based agents useful here is not raw language ability but the capacity to act in a loop: pose a query, read the result, decide whether to refine or move
   on \citep{wang2024survey}. We wire this loop to corpus infrastructure through the Model Context Protocol \citep[MCP;][]{mcp2024spec}. MCP presents each corpus      
  operation---frequency counts, collocation extraction, concordance retrieval---as a typed function the model can call mid-reasoning, so the agent never has to guess at
   statistics from its training data. Portability comes as a side benefit: because the protocol abstracts over the backend, the same agent transfers to any
  MCP-compatible corpus server.                                                                                                                                         
                  
  Every function call, together with its parameters and return value, is written to a structured log. This matters more than it might sound. Without such a trace, there
   is no way to tell whether a pattern the model reports was actually observed in the corpus or confabulated from pre-training.


\section{The Agent-Driven Framework}
\label{sec:framework}

This section presents the Agent-Driven approach to corpus linguistics, detailing its theoretical positioning, system architecture, and operational workflow.

\subsection{Definition and Positioning}
\label{sec:definition}

The distinction \citet{tognini2001corpus} drew between ``corpus-based'' and ``corpus-driven'' research still supplies the field's main methodological axis. \textit{Corpus-based} work starts from a hypothesis and seeks confirming or disconfirming evidence in the data; \textit{corpus-driven} work reverses the direction, letting patterns surface before committing to theory. The two differ in where theory enters, but both take for granted that a human analyst sits at the center of the loop---formulating queries, reading concordances, deciding what counts as a finding, and determining what to pursue next. Computational tools handle retrieval and counting; the interpretive and strategic work stays with the researcher.

We propose \textbf{Agent-Driven Corpus Linguistics}, in which the cognitive agent is a large language model (LLM) equipped with direct access to corpus query tools. We use ``agent'' in the sense established by \citet{wooldridge1995intelligent}: a system that exhibits \textit{autonomy} (operates without direct human intervention), \textit{reactivity} (responds to changes in its environment---here, corpus query results), and \textit{proactiveness} (takes initiative toward goals---here, generating and pursuing hypotheses). In this approach, the human researcher provides a research direction and exercises final editorial judgment, but the intermediate stages of hypothesis generation, query construction, result interpretation, hypothesis refinement, and discovery articulation are performed autonomously by the AI agent.

Crucially, the agent-driven approach is not a third position \textit{on the same axis} as the corpus-based/corpus-driven distinction. \citeauthor{tognini2001corpus}'s dichotomy concerns the \textit{epistemological direction} of inquiry: does theory precede data, or does data precede theory? The agent-driven dimension, by contrast, concerns the \textit{cognitive agent} performing the inquiry: is it a human or an AI? These two dimensions---the epistemological axis and the autonomy axis---are complementary rather than competing, and largely (though not fully) independent \citep[see][for extended discussion of this continuum]{hunston2002corpora, mcenery2012corpus}. An AI agent can operate in a corpus-based mode (testing hypotheses specified by the researcher), in a corpus-driven mode (letting patterns emerge from data without theoretical presupposition), or---as we demonstrate in this paper---in a hybrid mode where the agent generates its own hypotheses from parametric knowledge and then tests them against corpus evidence. In our case studies, the agent exhibited both modes: its initial hypotheses (H1--H4) reflected corpus-based reasoning from parametric knowledge, while the autonomous discovery of register sensitivity (H5) emerged in a corpus-driven manner from unexpected distributional patterns. What changes is not the epistemological relationship between theory and data, but \textit{who} navigates that relationship. We acknowledge that full orthogonality---a purely corpus-driven agent operating with no parametric priors---remains an idealization, since LLMs inevitably bring training-data knowledge to their analyses (see Section~\ref{sec:baseline} for further discussion). The agent, in other words, is not a faster pair of hands---it is an independent investigator.

It frames its own research questions, decides how to test them, and can surface patterns the researcher never thought to look for. At the same time, every finding produced by the agent is \textit{grounded} in actual corpus evidence---each conclusion can be traced to a specific query, a specific frequency count, a specific concordance line. A controlled experiment comparing our agent's output with the same LLM operating \textit{without} corpus access (Section~\ref{sec:baseline}) reveals that the grounding constraint provides three distinct forms of value: \textit{quantification} (transforming vague directional claims into precise figures), \textit{falsifiability} (enabling unexpected data to revise hypotheses), and \textit{data-responsive synthesis} (producing analytical frameworks that recombine existing theoretical concepts in light of corpus evidence).

Table~\ref{tab:paradigm_comparison} summarizes the key distinctions along this complementary dimension.

\begin{table}[ht]
\centering
\caption{Two largely independent dimensions of corpus linguistics methodology. The epistemological axis (corpus-based vs.\ corpus-driven) concerns the direction of inquiry; the agent axis concerns who performs the inquiry. Agent-driven research can operate in either epistemological mode.}
\label{tab:paradigm_comparison}
\small
\begin{tabularx}{\textwidth}{lXX}
\toprule
\textbf{Dimension} & \textbf{Human-Driven} & \textbf{Agent-Driven} \\
\midrule
Cognitive agent & Human researcher & AI (LLM) \\
Epistemological mode & Corpus-based, corpus-driven, or hybrid & Same (determined by task design) \\
Hypothesis source & Researcher's domain knowledge & Agent's parametric knowledge + data \\
Query construction & Manual & Autonomous \\
Result interpretation & Human analysis & AI analysis, human review \\
Iteration speed & Human-paced & Machine-paced \\
Hypothesis space & Shaped by researcher's expertise & Broader (systematic coverage) \\
Grounding & Corpus evidence & Corpus evidence \\
Risk of training-data bias & Low (human expertise) & Present (mitigated by grounding) \\
\bottomrule
\end{tabularx}
\end{table}

Three properties of the agent-driven approach merit emphasis. \textbf{Grounding} ensures that the agent's empirical claims are anchored in actual corpus data---each conclusion traces to a specific query, frequency count, or concordance line---while its interpretations remain LLM inferences subject to human evaluation (see Section~\ref{sec:baseline} for an empirical analysis of what grounding contributes). What grounding does not provide, however, is the capacity to follow up on a finding. This is where \textbf{iterative autonomy} becomes essential: the agent can notice an unexpected distributional pattern, hypothesise a cause, and  construct a targeted follow-up query---all within a single research session, without requiring the researcher to intervene between steps. Iterative autonomy, in turn, creates the conditions for \textbf{expanded analytical coverage}. Because the agent operates at machine speed, it can cross-tabulate findings against every available metadata dimension (genre, period, sentiment) as a matter of course.
 For instance, a researcher focused on diachronic intensifier change may reasonably choose not to cross-tabulate against all available genre categories in a single study; the agent, operating at machine speed, can explore all available metadata dimensions as a routine step. The advantage is not that the agent generates hypotheses beyond human capacity---the agent's five hypotheses in our case study are well within the scope of expert knowledge---but that it can provide \textit{systematic coverage at speed}, complementing the depth and theoretical sophistication that human expertise contributes.

\subsection{System Architecture}
\label{sec:architecture}

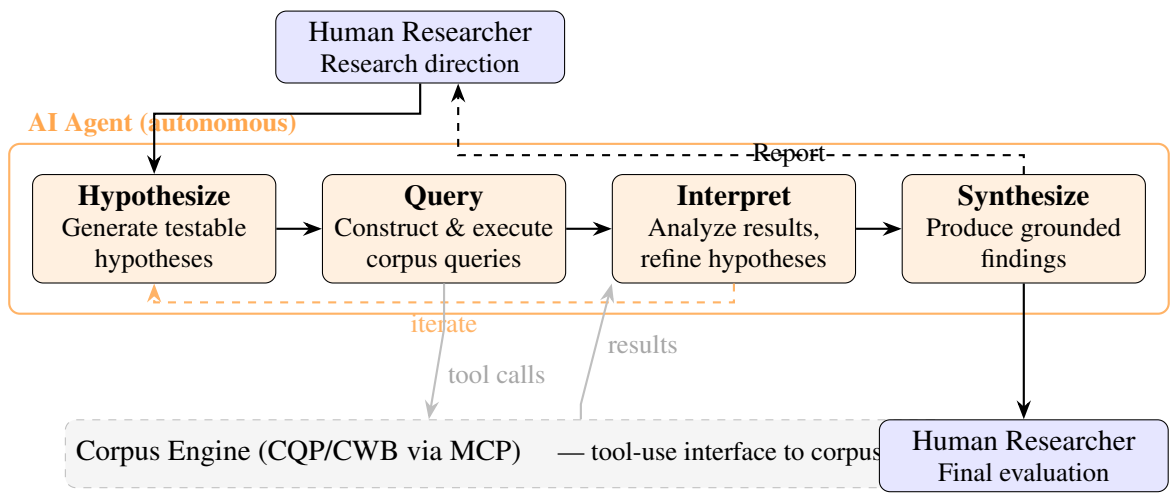
\begin{figure}[ht]
\centering
\begin{tikzpicture}[
  node distance=0.6cm and 1.5cm,
  humanbox/.style={draw, rounded corners, minimum width=3.8cm, minimum height=0.9cm, align=center, font=\small, fill=blue!10},
  agentbox/.style={draw, rounded corners, minimum width=3.2cm, minimum height=0.9cm, align=center, font=\small, fill=orange!12},
  techbox/.style={draw, rounded corners, minimum width=8cm, minimum height=0.9cm, align=center, font=\small, fill=gray!8, draw=gray!50, dashed},
  arrow/.style={-{Stealth[length=2.5mm]}, thick},
  looparrow/.style={-{Stealth[length=2.5mm]}, thick, dashed},
  label/.style={font=\footnotesize, text=gray!70}
]

\node[humanbox] (human) {Human Researcher\\[-2pt]{\footnotesize Research direction}};

\node[agentbox, below=1.2cm of human, xshift=-3.5cm] (hyp) {\textbf{Hypothesize}\\[-2pt]{\footnotesize Generate testable}\\[-2pt]{\footnotesize hypotheses}};
\node[agentbox, right=0.6cm of hyp] (query) {\textbf{Query}\\[-2pt]{\footnotesize Construct \& execute}\\[-2pt]{\footnotesize corpus queries}};
\node[agentbox, right=0.6cm of query] (interp) {\textbf{Interpret}\\[-2pt]{\footnotesize Analyze results,}\\[-2pt]{\footnotesize refine hypotheses}};
\node[agentbox, right=0.6cm of interp] (synth) {\textbf{Synthesize}\\[-2pt]{\footnotesize Produce grounded}\\[-2pt]{\footnotesize findings}};

\draw[draw=orange!60, rounded corners=4pt, thick] ([xshift=-0.3cm, yshift=0.4cm]hyp.north west) rectangle ([xshift=0.3cm, yshift=-0.4cm]synth.south east);
\node[font=\footnotesize\bfseries, text=orange!70, anchor=south west] at ([xshift=-0.2cm, yshift=0.35cm]hyp.north west) {AI Agent (autonomous)};

\draw[arrow] (hyp) -- (query);
\draw[arrow] (query) -- (interp);
\draw[arrow] (interp) -- (synth);

\draw[looparrow, orange!60] (interp.south) -- ++(0,-0.25) -| node[below, pos=0.25, font=\footnotesize, text=orange!60] {iterate} (hyp.south);

\draw[arrow] (human.south) -- ++(0,-0.4) -| (hyp.north);

\draw[arrow, dashed] (synth.north) -- ++(0,0.25) -| node[right, near start, font=\footnotesize] {Report} ([xshift=0.5cm]human.south);

\node[techbox, below=1.8cm of query, xshift=0.8cm] (tech) {Corpus Engine (CQP/CWB via MCP) \quad{\footnotesize --- tool-use interface to corpus data}};

\draw[arrow, gray!50] (query.south) -- ++(0,-0.6) -- node[right, label] {tool calls} ([xshift=-1cm]tech.north);
\draw[arrow, gray!50] ([xshift=1cm]tech.north) -- ++(0,0.2) -- node[right, label] {results} (interp.south west);

\node[humanbox, below=1.8cm of synth] (review) {Human Researcher\\[-2pt]{\footnotesize Final evaluation}};
\draw[arrow] (synth.south) -- node[right, font=\footnotesize] {} (review.north);

\end{tikzpicture}
\caption{System architecture of the agent-driven corpus linguistics framework. The orange loop at the centre is the agent's research cycle: hypothesize, query, interpret, revise. The corpus engine (gray, dashed) feeds grounded evidence into this loop through a tool-use interface; it is an implementation detail, swappable for any MCP-compatible backend. The human researcher sets the direction and has the  final word.}
\label{fig:architecture}
\end{figure}

The agent-driven framework is realized through a three-layer architecture connecting the AI agent to a remote corpus query engine via a standardized tool-use protocol. Figure~\ref{fig:architecture} illustrates the system design.

\paragraph{Layer 1: AI Agent.}
  The top layer is a large language model running inside an interactive coding environment. The researcher states a research direction in natural language; the agent
  takes it from there. It proposes an initial hypothesis, writes a corpus query to test it, reads the result, and decides whether to refine the hypothesis or branch 
  out---looping until it reaches a coherent set of findings. All corpus interaction goes through structured tool calls rather than free-form generation, so every
  retrieval step is logged with its exact parameters and can be reproduced.
                                                                                 
  \paragraph{Layer 2: CQP MCP Server.}                                               
  The middle layer is a lightweight Python server built on the Model Context Protocol \citep[MCP;][]{mcp2024spec}. It exposes three tools:
  \begin{itemize}[nosep]                                                               
  \item \texttt{corpus\_info()} reports what the agent needs before it can write its first query: field mappings, structural attributes, and the CQP syntax specific to
  the target corpus.                                                                    
  \item \texttt{cqp\_query(query, ...)} takes a CQP expression and returns concordance lines in KWIC format. The caller can narrow results by metadata fields and page
  through large result sets.                                                                
  \item The third, \texttt{cqp\_frequency(query, ...)}, is the quantitative counterpart: same query interface, but frequency counts rather than individual lines,
  optionally broken down by genre, author, or period.                                   
  \end{itemize}                                                                                                                                                                                                        
  We chose MCP for its standardization and built-in logging, but the framework does not depend on it. Any tool-use mechanism with structured function calling---OpenAI-style function calling, LangChain tool interfaces,  
  custom REST APIs---would work, as long as each invocation is a named-parameter function call whose inputs and outputs can be replayed. This is what makes the agent's empirical work auditable: a researcher can walk    
  through the exact sequence of tool calls and check every result against the corpus. On the server side, tool calls are translated into CQP command-line syntax, so the agent never writes raw CQP and the risk of        
  syntactic errors stays low.                                                                                                                                                                                              
                  
  \paragraph{Layer 3: CQP/CWB Engine.}                                                                                                                                                                                     
  The bottom layer is the IMS Open Corpus Workbench \citep[CWB;][]{evert2011cwb}. We work with a subset of the Standardized Project Gutenberg Corpus \citep[SPGC;][]{gerlach2020spgc}---69 English-language texts from
  antiquity to the late twentieth century, totalling roughly 5 million tokens. Annotation covers four token-level attributes (word form, lemma, POS tag, dependency relation) and two tiers of structural metadata:        
  text-level fields (author, genre category, publication period) and sentence-level fields (clause type, sentiment, subjectivity).

The connection between the tool-use server and the CQP engine supports multiple queries within a single research session, enabling the rapid query-interpret-iterate cycles that are central to the agent-driven workflow.

\subsection{Workflow}
\label{sec:workflow}

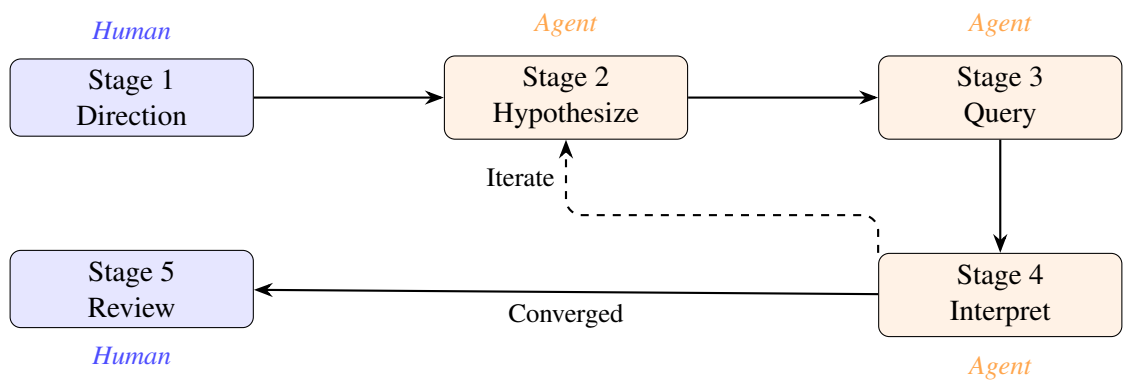
\begin{figure}[ht]
\centering
\begin{tikzpicture}[
  node distance=0.8cm and 2.5cm,
  stage/.style={draw, rounded corners, minimum width=3.2cm, minimum height=0.9cm, align=center, font=\small},
  arrow/.style={-{Stealth[length=2.5mm]}, thick},
  loopback/.style={-{Stealth[length=2.5mm]}, thick, dashed}
]

\node[stage, fill=blue!10] (s1) {Stage 1\\Direction};
\node[stage, fill=orange!10, right=of s1] (s2) {Stage 2\\Hypothesize};
\node[stage, fill=orange!10, right=of s2] (s3) {Stage 3\\Query};
\node[stage, fill=orange!10, below=1.5cm of s3] (s4) {Stage 4\\Interpret};
\node[stage, fill=blue!10, below=1.5cm of s1] (s5) {Stage 5\\Review};

\draw[arrow] (s1) -- (s2);
\draw[arrow] (s2) -- (s3);
\draw[arrow] (s3) -- (s4);
 \draw[loopback, rounded corners=6pt]
    (s4.north west) -- ++(0, 0.5) coordinate (itertop) -| (s2.south);      
  \node[font=\footnotesize, left] at ($(s2.south)!0.5!(s2.south |- itertop)$) {Iterate};
\draw[arrow] (s4) -- node[below, font=\footnotesize] {Converged} (s5);

\node[above=0.1cm of s1, font=\footnotesize\itshape, text=blue!70] {Human};
\node[above=0.1cm of s2, font=\footnotesize\itshape, text=orange!70] {Agent};
\node[above=0.1cm of s3, font=\footnotesize\itshape, text=orange!70] {Agent};
\node[below=0.1cm of s4, font=\footnotesize\itshape, text=orange!70] {Agent};
\node[below=0.1cm of s5, font=\footnotesize\itshape, text=blue!70] {Human};

\end{tikzpicture}
\caption{Five-stage iterative workflow. Blue stages involve human participation; orange stages are autonomous agent operations. The dashed arrow indicates the iteration loop.}
\label{fig:workflow}
\end{figure}

The agent-driven research process follows a five-stage iterative workflow, depicted in Figure~\ref{fig:workflow}.

\paragraph{Stage 1: Direction (Human).}                                                                                         
  The human researcher provides a research direction in natural language. This could be a broad prompt---``Investigate the use of English intensifiers in this corpus''---or something far more targeted: ``Compare the    
  collocational profiles of \textit{utterly} and \textit{terribly} across historical periods.'' Either way, the direction scopes the inquiry but leaves hypotheses and methods to the agent.                               
   
  \paragraph{Stage 2: Hypothesize (Agent).}                                                                                                
  The agent breaks the research direction into testable hypotheses. Given the intensifier prompt, for instance, it might independently propose hypotheses about frequency hierarchies, diachronic relay patterns, register
  sensitivity, and semantic delexicalization---four distinct lines of inquiry that a human researcher would typically spread across separate studies. For each hypothesis the agent also determines which corpus queries   
  would be needed to evaluate it.
                                                                                             
  \paragraph{Stage 3: Query (Agent).}                                                                                                 
  The agent constructs and executes CQP queries through the MCP tools, choosing query patterns, metadata filters, and frequency groupings suited to the hypothesis at hand. Queries are often chained: the result of one
  informs the design of the next.                                                                                                  
                  
  \paragraph{Stage 4: Interpret (Agent).}                                                                                                                    
  The agent reads the query results against its standing hypotheses. A hypothesis that survives testing stands; one that fits the data only partially gets narrowed; one that the evidence contradicts is dropped in favour
   of whatever pattern the data actually suggest. When the agent judges that further evidence is needed, it loops back to Stage~2 with revised hypotheses. The cycle repeats until the agent considers the evidence base   
  sufficient.     
                                                                                                    
  The loop between Stages~2--4 is what separates this workflow from fixed-pipeline systems such as UDagent \citep{lu2025udagent}, where the human specifies a question and the system executes a predetermined sequence of 
  operations. This is where genuine \textit{abductive reasoning} enters: the agent hits an unexpected distributional pattern, rethinks its explanation, and goes back to the corpus with a query designed to test the new
  account. That is how the agent ends up reporting findings the researcher never asked for.

\paragraph{Stage 5: Review (Human).}                                                                                                                                  
  Once the agent's exploration converges, it hands the researcher a structured report---queries, frequency tables, interpretive reasoning, all included. What happens   
  next is up to the researcher: accept a finding, flag an artifact, strike a conclusion that does not survive scrutiny, or send the agent back to Stage~1 with a sharper
   question. The division of labour stays constant throughout---the researcher owns the questions and the final judgement, the agent owns the legwork.
\section{Case Studies}
\label{sec:case_study}

 We test the framework on English intensifiers---\textit{very}, \textit{really}, \textit{extremely}, \textit{utterly}. The domain suits our purpose: intensifiers
  change fast, move unevenly across registers, and have been studied closely enough that we know what the agent should recover. If it finds something the literature
  missed, we will know that too.

  \subsection{Intensifiers in Corpus Linguistics}
  \label{sec:intensifier_background}

The study of English intensifiers has a longer pedigree than corpus linguistics itself. \citet{stoffel1901intensives} catalogued ``intensives and down-toners'' over a century ago, and many of his distributional observations still hold. \citet{bolinger1972degree} supplied the typology most later work builds on;
  \citet{quirk1985comprehensive} refined it into the now-standard split between \textit{amplifiers} (maximizers and boosters) and \textit{downtoners}. Several strands of research have built on this foundation. One documents the recurring ``recycling'' of intensifiers---the   
  rise and fall of successive dominant forms over historical time---across multiple varieties of English \citep{tagliamonte2008intensifiers, ito2003well, tagliamonte2005so, bauer2002adjective}, while
  \citet{nevalainen2002fairly} traced the long-term grammaticalization trajectories of degree modifiers from Middle English onward. \citet{mendeznaya2003intensifiers} examined the grammaticalization of early English
  intensifiers, and \citet{lorenz2002really} investigated the process of delexicalization by which intensifiers lose their original semantic content and become pure degree modifiers. On the semantic side,
  \citet{paradis1997degree} demonstrated that degree modifiers are constrained by the scalar properties of the adjectives they modify, distinguishing \textit{scalar} from \textit{limit} adjectives;
  \citet{partington1993semantic} investigated collocational evidence for semantic change in intensifiers; and \citet{xiao2007corpus} provided a large-scale corpus-based study of amplifier distribution in British
  English. The grammaticalization processes underlying these changes have been theorized within broader frameworks of semantic change, including \citeauthor{traugott1982propositional}'s
  (\citeyear{traugott1982propositional}) model of subjectification (propositional $\to$ textual $\to$ expressive meaning) and the cline from content item to grammatical word described by
  \citet{hopper2003grammaticalization}. From a sociolinguistic perspective, \citet{labov1984intensity} established that intensifier choice is conditioned by social factors including age and gender.

  Three findings from this body of work matter most for our purposes. Intensifiers exhibit a life cycle of innovation, spread, and eventual decline as newer forms emerge \citep{tagliamonte2008intensifiers}.             
  Delexicalization proceeds at different rates and along different pathways depending on the form. And register and social factors---age, formality, genre---condition both adoption and distribution
  \citep{tagliamonte2008intensifiers, labov1984intensity}. Together, these results give us a known baseline: if the agent's autonomous analysis recovers them, we can trust its methodology; if it goes beyond them, we    
  have something new to evaluate.

\subsection{Research Setup}
\label{sec:setup}

 The corpus used in this study is a 5,019,103-token subset of the Standardized Project Gutenberg Corpus \citep{gerlach2020spgc}, consisting of 69
  English-language texts indexed with the IMS Open Corpus Workbench \citep{evert2011cwb}. Each token is annotated with word form, lemma, coarse and fine          
  part-of-speech tags, dependency relation, and head word. At the text level, structural attributes cover author, genre category (13 categories, from
  \textit{Poetry} to \textit{Law/Criminology}), and historical period. Sentence-level attributes include clause type, sentiment, and subjectivity.                
                                                                                                            
  To support diachronic analysis, we added temporal metadata. We obtained publication or composition years from the Gutendex API, which provides author birth and 
  death years, and from dates embedded in text titles (e.g., ``The Atlantic Monthly, July 1858''). These years were then grouped into eight historical periods
  from \textit{ancient} (BCE) to \textit{late 20th century} (1950+), covering 66 of the 69 texts.                                                                 
                                                                                                       
  The AI agent was Anthropic's Claude Opus, a large language model accessed through a command-line coding environment with default decoding parameters            
  (temperature = 1.0; no nucleus sampling or top-$k$ truncation). We chose this model for its reliable tool-use capabilities and its ability to maintain coherence
   over the long multi-step investigations that our workflow demands. The sole human instruction was:                                                             
  \begin{quote}                                                                                                                                                 
  ``Investigate the use of English intensifiers in this corpus.''
  \end{quote}                                                                                                                                                     
  Everything that follows---the hypotheses, queries, analytical methods, and findings---was produced autonomously by the agent.

\subsection{Agent Exploration Process}
\label{sec:exploration}

The agent's investigation proceeded through five rounds of iterative exploration, corresponding to the workflow described in Section~\ref{sec:workflow}.

\paragraph{Round 1: Hypothesis generation.}
  Without further prompting, the agent arrived at four hypotheses:

  \begin{itemize}[nosep]
    \item[\textbf{H1}:] Intensifiers follow a ``relay'' pattern: as one form declines, another rises to take its place.
    \item[\textbf{H2}:] Classical forms (\textit{exceedingly}, \textit{vastly}) now occur at far lower frequencies than their modern replacements (\textit{very},
  \textit{really}).
    \item[\textbf{H3}:] Certain intensifiers have bleached semantically---they have lost their original lexical content and function as pure degree markers.
    \item[\textbf{H4}:] Genre matters: different intensifiers cluster in different text types.
  \end{itemize}
  The agent drew on its parametric knowledge of historical linguistics and corpus methodology---identifying, on its own, four productive lines of inquiry that a human researcher might well have chosen but did not specify.

\paragraph{Round 2: Frequency profiling.}
The agent queried the corpus for 22 intensifiers, drawing on a set broadly consistent with the amplifier categories identified by \citet{quirk1985comprehensive}.\footnote{The 22 intensifiers queried were: \textit{very}, \textit{so}+ADJ, \textit{most}+ADJ, \textit{rather}, \textit{quite}, \textit{really}, \textit{extremely}, \textit{truly}, \textit{highly}, \textit{absolutely}, \textit{entirely}, \textit{utterly}, \textit{perfectly}, \textit{deeply}, \textit{greatly}, \textit{remarkably}, \textit{dreadfully}, \textit{exceedingly}, \textit{terribly}, \textit{awfully}, \textit{frightfully}, and \textit{thoroughly}. To address potential ambiguity, \textit{so}+ADJ was restricted to the pattern \texttt{[word="so"\%c] [pos="JJ.*"]}, excluding resultative and conjunctive uses; \textit{most}+ADJ was queried as \texttt{[word="most"\%c] [pos="JJ.*"]} and manually filtered to exclude superlative constructions (e.g., \textit{the most beautiful}) by requiring the absence of a preceding determiner.} The frequency results revealed a clear core--periphery structure: \textit{very} (1,089 per million words, henceforth pmw) dominated, followed by \textit{so}+ADJ (703~pmw), \textit{most}+ADJ (540~pmw), \textit{rather} (258~pmw), \textit{quite} (243~pmw), and \textit{really} (174~pmw), with forms such as \textit{terribly} (10~pmw), \textit{awfully} (4~pmw), and \textit{frightfully} (2~pmw) at the margins. The agent noted:

\begin{quote}
\textit{``The frequency data reveal a clear core--periphery layering: very, so+ADJ, and most+ADJ form the core ($>$500 pmw), while terribly, awfully, and frightfully are marginal ($<$20 pmw). This gradient is consistent with H2 and suggests that intensifier replacement is not abrupt but proceeds through intermediate stages.''}
\end{quote}

\paragraph{Round 3: Autonomous discovery of register sensitivity.}
When the agent cross-tabulated intensifier frequencies against text genre, it independently identified a pattern not anticipated by the researcher. The agent observed:

\begin{quote}
\textit{``The normalized frequency of really in dramatic texts (352~pmw) is strikingly higher than in poetry (17~pmw)---a 20-fold difference. This suggests that really functions as a colloquial marker whose literary occurrence is concentrated in dialogue-simulating registers. I propose a new hypothesis H5: the genre distribution of intensifiers reflects their position on a formality continuum.''}
\end{quote}

This autonomous generation of H5 is worth unpacking. Register variation in intensifiers is not unknown in the literature \citep{tagliamonte2008intensifiers,
  xiao2007corpus}, but it was not part of the research directive, nor was it prompted by the human researcher. What happened is that the agent noticed a specific 
  quantitative pattern (the 20-fold difference between \textit{really} in dramatic texts and poetry), recognised it as meaningful, connected it to a broader
  framework (register variation as a formality continuum), and turned it into a testable hypothesis---all on its own. The point is not that register sensitivity  
  is a new idea, but that the agent found its way to it through the data without being pointed in that direction.

\paragraph{Round 4: Semantic prosody quantification.}
  For H3---semantic delexicalization---the agent moved beyond frequency data and devised its own quantification scheme, unprompted. It extracted all adjective
  collocates of each intensifier from the pattern \texttt{[word="\textit{X}"\%c] [deprel="ADJ"]} and computed three metrics. Collocational diversity measured the number
   of distinct adjective types and Top-5 concentration; the hapax ratio captured how many collocates appeared only once; and collocational polarity
  \citep[cf.][]{stubbs2001words} sorted each collocate into one of four categories---positive, negative, neutral, or privative (\textit{impossible}, \textit{unknown}).
  The agent reported:

  \begin{quote}
  \textit{``Utterly exhibits a striking negative polarity: 80\% of its collocates are negative or privative (impossible, unknown, unable, useless, hopeless...). This is
   not delexicalization---it is semantic narrowing. The original meaning `to the outermost' has specialized into marking total absence or negation.''}
  \end{quote}

\paragraph{Round 5: Diachronic analysis.}
 In the final round, the agent used the newly added temporal metadata to track frequency trends across eight historical periods. This allowed it to test H1 (the relay hypothesis) directly with diachronic evidence, going beyond the synchronic distributional data of earlier rounds.

\subsection{Findings}
\label{sec:findings}

We organize the findings by the hypotheses that structured the agent's exploration, illustrating how each was tested, refined, or extended through iterative corpus inquiry.

\subsubsection{The Intensifier Relay (H1)}
\label{sec:relay}

The diachronic frequency data provide suggestive evidence consistent with the relay hypothesis. Table~\ref{tab:diachronic} presents normalized frequencies (pmw) across historical periods.

\begin{table}[ht]
\centering
\caption{Normalized frequency (per million words) of selected intensifiers across historical periods.}
\label{tab:diachronic}
\small
\begin{tabular}{lrrrrrrrr}
\toprule
& \textbf{anc.} & \textbf{med.} & \textbf{e.mod.} & \textbf{18C} & \textbf{e.19C} & \textbf{l.19C} & \textbf{e.20C} & \textbf{l.20C} \\
\midrule
\textit{very}     & 472  & 576   & 631   & \textbf{1,804} & 1,287 & 1,281 & 959  & 964  \\
\textit{so}+ADJ   & 459  & \textbf{966}  & \textbf{1,017} & 712   & 823   & 705   & 629  & 418  \\
\textit{quite}    & 179  & 76    & 159   & 225   & 222   & 265   & 211  & \textbf{431}  \\
\textit{really}   & 77   & 54    & 40    & 151   & 42    & 199   & 190  & 185  \\
\textit{truly}    & 0    & 54    & 99    & \textbf{170}   & 84    & 106   & 64   & 50   \\
\textit{utterly}  & 13   & \textbf{101}  & 20    & 18    & 74    & 31    & 29   & 48   \\
\bottomrule
\end{tabular}
\\[4pt]
\footnotesize anc.=ancient; med.=medieval; e.mod.=early modern; e./l.=early/late. Bold indicates period peak.
\end{table}

The frequency data point to three relay chains (though the small corpus makes period-level figures approximate):
  \begin{enumerate}[nosep]
  \item \textbf{\textit{so}+ADJ $\to$ \textit{very}}: \textit{So}+ADJ dominates the medieval and early modern periods (966--1,017~pmw) but falls to 418~pmw by the late
  20th century---a 59\% drop. \textit{Very} fills the gap, peaking at 1,804~pmw in the 18th century before levelling off around 960~pmw.
  \item \textbf{\textit{truly} $\to$ \textit{really}}: \textit{Truly} peaks in the 18th century at 170~pmw and fades to 50~pmw. \textit{Really} trends upward
  overall---57~pmw early on to 187~pmw late, a 3.3-fold increase---but the trajectory is anything but smooth: 151~pmw in the 18th century, down to 42~pmw in the early
  19th, back up to 199~pmw by the late 19th. The most likely culprit is the small number of texts per period and their uneven register mix, not a genuine reversal.
  \item \textbf{\textit{utterly} contraction}: \textit{Utterly} starts high in the medieval period (101~pmw) and shrinks steadily---a trajectory consistent with the
  semantic narrowing discussed below.
  \end{enumerate}

Log-likelihood tests indicate that the diachronic shifts for \textit{very} ($G^2 = 107.1$, $p < 0.0001$, Cram\'{e}r's $V = 0.20$), \textit{so}+ADJ ($G^2 = 68.0$, $p < 0.0001$, $V = 0.16$), and \textit{really} ($G^2 = 69.1$, $p < 0.0001$, $V = 0.16$) are statistically significant when comparing the early period (ancient through early modern; 697,053 tokens) against the late period (early--late 20th century; 2,045,266 tokens). Effect sizes range from small to medium. The changes for \textit{truly} and \textit{utterly} do not reach significance ($G^2 < 1$, n.s.), likely due to their low absolute frequencies.

The relay chains echo what \citet{tagliamonte2008intensifiers} and \citet{ito2003well} found in spoken Canadian and British English, though their data came from
  contemporary speech rather than centuries of literary prose. Our corpus is too small to be definitive on period-level trends, but the broad pattern---successive
  displacement, not parallel coexistence---holds across both spoken and written registers and over a considerably longer window than either study covered. Larger,
  temporally balanced corpora would be the obvious next test.

\subsubsection{Semantic Change Trajectories (H3)}
\label{sec:delexicalization}

Collocational data point to three distinct pathways of semantic change, not the single trajectory toward delexicalization that a textbook account might suggest. Each
  pathway has been described in the grammaticalization literature under a different heading---desemanticization \citep{lehmann2015thoughts,
  heine2003grammaticalization}, semantic narrowing through invited inferencing \citep{traugott2002regularity}, and metaphorical extension
  \citep{sweetser1990etymology}---but the three have not, to our knowledge, been quantified side by side within a single intensifier system:

  \paragraph{Pathway 1: Complete delexicalization.}
  \textit{Very} is the clearest case. Its Top-5 concentration is the lowest in the dataset (12.4\%), it modifies 807 distinct adjective types, and its collocate profile
   splits almost evenly across polarities (44\% positive, 44\% neutral, 12\% negative). Nothing in the distribution picks out a preferred semantic domain---the word has
   become a pure degree marker. The original meaning (from Latin \textit{verus}, ``true'') has been entirely bleached; \textit{very} functions as a pure degree amplifier.

\paragraph{Pathway 2: Polarity fixation (subjectification).}
\textit{Utterly} and \textit{truly} illustrate a contrasting pathway in which delexicalization is arrested by polarity fixation---the specialization of an intensifier toward collocates of a particular evaluative polarity. \textit{Utterly}'s collocates are overwhelmingly negative or privative (80\%): \textit{impossible}, \textit{unknown}, \textit{unable}, \textit{useless}, \textit{hopeless}, \textit{worthless}, \textit{meaningless}, \textit{incomprehensible}, \textit{devoid}. The etymological meaning (``to the outer extreme'') has not been lost but has been channeled into a specialized semantic niche---the intensification of absence, failure, or negation. Conversely, \textit{truly} retains a strong positive bias (68\% positive collocates: \textit{great}, \textit{beautiful}, \textit{glorious}, \textit{magnificent}, \textit{wonderful}), preserving its connection to \textit{truth} and authenticity.

We note that within the negative/privative category, the agent distinguished \textit{privative} collocates---those denoting absence, lack, or negation (e.g., \textit{impossible}, \textit{unknown}, \textit{devoid})---from general \textit{negative} collocates expressing negative evaluation without the component of absence (e.g., \textit{useless}, \textit{hopeless}). This distinction is semantic rather than morphological: \textit{dead} is classified as negative (an undesirable state), while \textit{impossible} is privative (the absence of possibility). The privative sub-category proved analytically productive because it captures \textit{utterly}'s specialization toward total absence or negation---a pattern obscured if privative items are merged with general negative evaluations.

The trajectory of \textit{utterly} cuts against the conventional story of unidirectional delexicalization. In \citeauthor{traugott1982propositional}'s
  (\citeyear{traugott1982propositional}) terms, the word has subjectified: its propositional core (spatial extremity) gave way to an expressive-evaluative function (the
   speaker's assessment of completeness), but the process stopped short of full bleaching. \textit{Utterly} narrowed into a specific evaluative niche rather than
  generalizing across the board---semantic specialization, not desemanticization \citep[cf.][]{hopper2003grammaticalization}.

\paragraph{Pathway 3: Metaphorical constraint.}
  \textit{Deeply} occupies a third position. Its top collocates---\textit{pinnatifid} (8~occurrences), \textit{rooted} (8), \textit{impressed} (4), \textit{indebted}
  (4), \textit{interested} (3)---are revealing: four of the five still trade on the spatial metaphor of depth, whether literally (\textit{pinnatifid}, \textit{rooted})
  or via a psychological extension (\textit{impressed}, originally ``pressed into'').\footnote{The high count for \textit{pinnatifid} traces to a single botanical text
  (Darwin's \textit{Variation of Animals and Plants Under Domestication}). This is a corpus-composition artifact, but it makes a useful point: \textit{deeply} can still
   collocate with its literal spatial sense---something \textit{very} lost long ago.} The original source domain---physical depth---has not let go; it still filters
  which adjectives \textit{deeply} can intensify, exactly the kind of scalar sensitivity \citeauthor{paradis1997degree} (\citeyear{paradis1997degree}) predicted. The
  result is a narrow collocational profile (Top-5 concentration: 54\%). \textit{Deeply} has stretched metaphorically, but it has not bleached.

\begin{figure}[ht]
\centering
\begin{tikzpicture}[
  font=\small,
  word/.style={font=\small\itshape},
  label/.style={font=\scriptsize, text=gray}
]
\draw[-{Stealth}, thick] (0,0) -- (13,0);
\node[above] at (0,0.3) {\textbf{Fully delexicalized}};
\node[above] at (13,0.3) {\textbf{Source meaning retained}};

\node[word, below] at (0.5,-0.3) {very};
\node[label, below] at (0.5,-0.8) {neutral};

\node[word, below] at (2.5,-0.3) {quite};
\node[label, below] at (2.5,-0.8) {neutral};

\node[word, below] at (4.5,-0.3) {rather};
\node[label, below] at (4.5,-0.8) {neutral};

\node[word, below] at (6.5,-0.3) {really};
\node[label, below] at (6.5,-0.8) {mixed};

\node[word, below] at (8.5,-0.3) {truly};
\node[label, below] at (8.5,-0.8) {positive};

\node[word, below] at (10.5,-0.3) {terribly};
\node[label, below] at (10.5,-0.8) {negative};

\node[word, below] at (12.5,-0.3) {utterly};
\node[label, below] at (12.5,-0.8) {neg+privative};

\foreach \x in {0.5,2.5,4.5,6.5,8.5,10.5,12.5}
  \fill (\x, 0) circle (3pt);

\end{tikzpicture}
\caption{Semantic change trajectories of English intensifiers, arranged by collocational freedom (left = maximal; right = constrained). The horizontal axis represents collocational diversity, not a single underlying continuum: complete delexicalization, polarity fixation (subjectification), and metaphorical constraint are qualitatively distinct processes \citep{hopper2003grammaticalization, traugott1982propositional, sweetser1990etymology} that produce different collocational profiles.}
\label{fig:spectrum}
\end{figure}
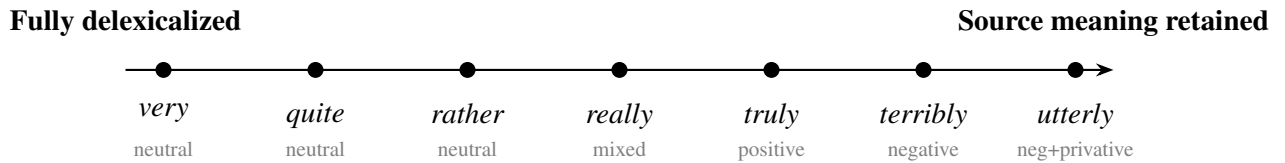

Figure~\ref{fig:spectrum} summarizes the delexicalization spectrum.

\subsubsection{Register Sensitivity and Colloquialization (H4 + H5)}
\label{sec:register}

The agent's discovery of register-dependent intensifier distribution (H5) was confirmed by systematic cross-tabulation. Table~\ref{tab:register} presents normalized frequencies for selected intensifiers across genre categories.

\begin{table}[ht]
\centering
\caption{Normalized frequency (pmw) of intensifiers in selected genres.}
\label{tab:register}
\small
\begin{tabular}{lrrrr}
\toprule
& \textbf{Drama} & \textbf{Poetry} & \textbf{Journalism} & \textbf{Cooking} \\
\midrule
\textit{very}      & 657  & 635  & 1,619 & 2,160 \\
\textit{really}    & 352  & 17   & 162   & 119   \\
\textit{so}+ADJ    & 800  & 919  & 742   & 340   \\
\textit{quite}     & 219  & 146  & 289   & 85    \\
\textit{exceedingly} & ---  & 8    & 38    & ---   \\
\bottomrule
\end{tabular}
\end{table}

The data sort intensifiers into three register profiles:
  \begin{itemize}[nosep]
  \item \textit{Really} is a \textbf{colloquial marker}. In dramatic texts it hits 352~pmw---twenty times its frequency in poetry (17~pmw; $G^2 = 79.0$, $p < 0.0001$, $V = 0.41$), the largest register gap in the dataset.
  \item \textit{Very} looks \textbf{register-neutral} across literary genres but spikes in utilitarian prose (Cooking: 2,160~pmw; Journalism: 1,619~pmw), well above
  poetry (635~pmw; $G^2 = 104.5$, $p < 0.0001$, $V = 0.19$). It appears to have settled into the role of default intensifier for functional communication.
  \item \textbf{Cross-register} \textit{so}+ADJ shows a poetic tilt (919~pmw in Poetry), perhaps because of its natural fit with exclamatory constructions.
  \end{itemize}
  These register patterns dovetail with the diachronic picture. \textit{Really} is both the fastest-rising intensifier (3.3-fold increase) and the most
  register-skewed---a combination that fits the broader colloquialization of written English documented by \citet{lorenz2002really}.

\subsubsection{Synthesis}
\label{sec:synthesis}

The agent's five rounds of exploration produced findings across four dimensions: frequency hierarchy, diachronic trajectory, semantic prosody, and register distribution. Crucially, these dimensions are not independent. The diachronic relay (\textit{so}+ADJ $\to$ \textit{very} $\to$ \textit{really}) intersects with register (\textit{really} rises because colloquial forms gain ground in written texts), with delexicalization (\textit{very} is fully delexicalized and therefore register-neutral, while \textit{utterly} retains semantic constraints that limit its use), and with collocational patterns (fully delexicalized forms show high collocational diversity; polarity-fixed forms show restricted profiles). The capacity to identify these cross-dimensional connections autonomously---without human prompting---is a distinctive affordance of the agent-driven approach.

\subsection{Validation Against Existing Literature}
\label{sec:validation}

 We can check these findings against a well-established literature:
  \begin{itemize}
  \item \textbf{Relay pattern}: The agent traced the same successional ``recycling'' that \citet{tagliamonte2008intensifiers} and \citet{ito2003well} documented in
  spoken English---but in written literary texts spanning the medieval period to the late 20th century, a considerably longer window.
  \item \textbf{Delexicalization}: That intensifiers follow multiple evolutionary pathways, not a single bleaching trajectory, echoes \citet{lorenz2002really}. What the
   agent adds is the identification of polarity fixation as a pathway in its own right, distinct from an intermediate stage of bleaching.
  \item \textbf{Register sensitivity}: \citet{tagliamonte2008intensifiers} tied \textit{really} to younger speakers and informal contexts in spoken corpora. The 20-fold
   gap between dramatic and poetic texts in our literary corpus tells a parallel story from a different modality.
  \item \textbf{Semantic prosody}: The concept of semantic prosody, introduced by \citet{louw1993irony} and \citet{sinclair1991corpus}, was independently
  operationalized by the agent through its four-way collocate classification. The finding that \textit{utterly} carries a strong negative/privative prosody is
  consistent with established observations \citep{partington1993semantic}, though the agent's quantification (80\% negative+privative collocates) provides more precise
  evidence than prior qualitative accounts.
  \end{itemize}

In summary, the agent's findings are broadly consistent with the existing literature while contributing several novel observations: the three-pathway model of delexicalization, the quantified polarity profiles, and the cross-dimensional synthesis linking diachronic, register, and semantic evidence.

\subsection{What Does Corpus Grounding Add?}
\label{sec:baseline}

A fundamental challenge for agent-driven research is distinguishing genuine \textit{discovery} from the corpus data from \textit{recall} of patterns already present in the LLM's training data. To address this, we conducted a controlled experiment: we presented the same research directive (``Investigate English intensifiers in literature spanning from antiquity to the 20th century'') to the \textit{same} LLM (Claude Opus) with \textit{no} corpus access, asking it to provide its best analysis from training data alone. By holding the model constant and varying only the availability of corpus tools, this design isolates the contribution of empirical grounding from the model's parametric knowledge.

Table~\ref{tab:baseline} summarizes what the ungrounded model could and could not provide.

\begin{table}[ht]
\centering
\caption{Comparison of the same LLM (Claude Opus) with and without corpus access.}
\label{tab:baseline}
\small
\begin{tabularx}{\textwidth}{lXX}
\toprule
\textbf{Capability} & \textbf{Without corpus (recall)} & \textbf{With corpus (agent-driven)} \\
\midrule
Hypothesis generation & 6 well-formed hypotheses, consistent with literature & 5 hypotheses, including 1 data-prompted (H5) \\
Frequency hierarchy & Correct rank ordering, no numbers & Precise pmw for 22 intensifiers \\
Diachronic trends & Correct directional predictions & 8-period quantified data; \textit{very} peak at 1,804~pmw in 18C \\
Register sensitivity & Predicted informal $>$ formal & Quantified: \textit{really} 352 vs.\ 17~pmw (20$\times$) \\
Delexicalization & Described general multi-stage pathway & Three-pathway comparative framework with quantified polarity profiles \\
Collocate lists & Impressionistic examples only & 807 types for \textit{very}; 67 for \textit{utterly}; full classification \\
Specific numbers & Self-assessed as ``Very Low'' confidence & Every figure traceable to a CQP query \\
\bottomrule
\end{tabularx}
\end{table}

The experiment reveals that grounding provides value at three levels:

\begin{enumerate}[nosep]
  \item \textbf{Quantification}: The ungrounded LLM predicted that \textit{really} would be more frequent in informal registers, but could not quantify the magnitude. The grounded agent established the 20-fold difference (352 vs.\ 17~pmw), transforming a directional hypothesis into a precise empirical claim.
  \item \textbf{Falsifiability}: The ungrounded LLM predicted that ``collocational narrowing precedes bleaching.'' The grounded agent's data allowed this prediction to be tested against actual collocational profiles---and potentially revised.
  \item \textbf{Data-responsive synthesis}: The three-pathway comparative framework (complete delexicalization, polarity fixation, metaphorical constraint) was not produced by the ungrounded model. While the individual pathways correspond to processes described in the grammaticalization literature \citep{lehmann2015thoughts, traugott2002regularity, sweetser1990etymology}, the agent's quantified collocational comparison---which enabled simultaneous empirical differentiation of the three trajectories---arose from its systematic classification of collocate polarity profiles in response to the corpus data.
\end{enumerate}

The baseline experiment also clarifies what agent-driven research does \textit{not} add: the initial hypothesis generation was equally competent without corpus access, confirming that this step draws primarily on the LLM's training data rather than on empirical discovery. The agent-driven approach is thus best understood not as a way to generate hypotheses that humans cannot imagine, but as a way to \textit{empirically ground, quantify, and extend} hypotheses at machine speed---including the capacity to construct data-responsive analytical frameworks by combining existing theoretical concepts in light of quantified corpus evidence.

\subsection{Replicating Published Research}
\label{sec:replication}

The preceding sections demonstrate that the agent can autonomously explore a corpus and produce findings consistent with the existing literature. A stronger test is whether the framework can \textit{replicate} specific quantitative results from published studies---comparing its output not against qualitative expectations but against precise numbers reported by other researchers. We selected two studies conducted on the Corpus of Late Modern English Texts \citep[CLMET 3.1;][]{desmet2015clmet}, a 40.3-million-token corpus of 333 texts spanning 1710--1920. Crucially, neither study concerns intensifiers, providing an additional test of the framework's generalizability beyond the domain of the primary case study.

\subsubsection{Claridge (2025): Diachronic Decline of \textit{reader}}

\citet{claridge2025reader} documented the diachronic decline of the noun \textit{reader} across CLMET's three sub-periods, reporting a total frequency of 88.3~pmw and a decline from approximately 120~pmw (1710--1780) to 52.5~pmw (1850--1920). We instructed the agent to query \texttt{[word="reader"]} in CLMET and compute normalized frequencies per period.

\begin{table}[ht]
\centering
\caption{Replication of \citet{claridge2025reader}: normalized frequency (pmw) of \textit{reader} (singular) in CLMET.}
\label{tab:claridge_replication}
\small
\begin{tabular}{lrrr}
\toprule
& \textbf{1710--1780} & \textbf{1780--1850} & \textbf{1850--1920} \\
\midrule
Agent result & 105.6 & 90.7 & 53.3 \\
Claridge (2025) & $\sim$120 & $\sim$99 & $\sim$52.5 \\
\bottomrule
\end{tabular}
\end{table}

The declining trend is confirmed across all three periods. The third-period value is nearly identical (53.3 vs.\ 52.5~pmw). The small discrepancies in the earlier periods likely reflect differences in normalization base (our figures use all tokens including punctuation vs.\ Claridge's word-only count) and minor version differences between CLMET~3.0 and~3.1.

\subsubsection{De Smet (2013): Gerund Complement Spreading}

\citet{desmet2013spreading}---the creator of CLMET---documented the historical spread of gerund complements at the expense of to-infinitives. The verb \textit{remember} provides the clearest test case: De Smet showed that \textit{remember}+gerund (e.g., \textit{remember seeing}) gradually replaced \textit{remember}+to-infinitive (e.g., \textit{remember to see}) across the Late Modern period. We queried both patterns:

\begin{table}[ht]
\centering
\caption{Replication of \citet{desmet2013spreading}: \textit{remember}+gerund vs.\ \textit{remember}+to-infinitive in CLMET.}
\label{tab:desmet_replication}
\small
\begin{tabular}{lrrr}
\toprule
& \textbf{1710--1780} & \textbf{1780--1850} & \textbf{1850--1920} \\
\midrule
\textit{remember}+VBG & 4 & 61 & 162 \\
\textit{remember}+to-inf & 71 & 74 & 48 \\
Gerund share & 5.3\% & 45.2\% & \textbf{77.1\%} \\
\bottomrule
\end{tabular}
\end{table}

The gerund complement rises from 5.3\% to 77.1\% of all \textit{remember} complements, confirming De Smet's thesis of spreading diffusion. Five additional verbs tested (\textit{enjoy}, \textit{finish}, \textit{regret}, \textit{suggest}, \textit{consider}) show the same pattern, with gerund complements increasing across all three periods.

\subsubsection{Summary}

  Both replications land close to the published numbers. The Claridge study matches period-specific frequencies almost exactly; the De Smet study recovers a subtler
  pattern---the 15-fold rise in gerund complements---that might easily have been missed without iterative querying. That the two span different phenomena (lexical
  frequency vs.\ complementation syntax) is the stronger result: it suggests the framework is not tied to the intensifier domain.

\section{Discussion}
\label{sec:discussion}

\subsection{Evaluating the Agent-Driven Approach}

The case study highlights five ways the agent-driven approach extends what a conventional corpus study can achieve in a single pass.                                  
   
  First, \textbf{autonomous research from directive to findings.} A one-sentence directive---``investigate English intensifiers''---was enough. The agent generated five
   hypotheses, tested each against the corpus, revised its framework when the data pushed back, and delivered a synthesis spanning diachronic, semantic, and register
  dimensions, all without intermediate human input. In a traditional workflow every one of these steps sits on the researcher's desk; here, the researcher set a      
  direction and the agent ran the inquiry. Two sources fed the hypotheses: H1--H4 drew on the kind of parametric knowledge a human expert would also bring, while H5
  (register sensitivity) emerged from cross-tabulation and pushed the investigation past the original directive's scope. That the findings largely converge with
  established literature is expected---English intensifiers are well studied---but convergence validates the framework, it does not define its ceiling. An agent turned
  loose on a less-charted phenomenon may well surface patterns no human has yet catalogued.

  Second, \textbf{grounded discovery.} Every claim the agent makes is anchored to a specific corpus query. When it reports that \textit{utterly} carries an 80\%      
  negative/privative collocational profile, the figure traces back to an exhaustive classification of 67 collocate types extracted via a documented CQP query---not to a  plausible-sounding guess from training data.

  Third, \textbf{cross-dimensional synthesis.} The agent did not analyze diachronic, register, and semantic dimensions in isolation---it spotted connections between  them as the analysis iterated. \textit{Really}'s diachronic rise correlates with its concentration in colloquial registers; \textit{utterly}'s medieval frequency peak corresponds to its subsequent semantic narrowing.These integrative observations illustrate how the agent's capacity to hold multiple analytical threads within a single session can complement the focused, depth-first approach that characterizes much human-led corpus research.

Fourth, the \textbf{baseline experiment} (Section~\ref{sec:baseline})---which compared the same model (Claude Opus) with and without corpus access, thereby isolating the contribution of empirical grounding from the model's parametric knowledge---provides empirical evidence for a three-level typology of agent output:

\begin{itemize}[nosep]
  \item \textbf{Level 1: Recall.} Hypotheses and qualitative predictions that the LLM can produce from training data alone, without corpus access. Example: the prediction that intensifiers undergo diachronic replacement, or that \textit{utterly} has negative collocational preferences. These outputs reflect the LLM's parametric knowledge and are not contributions of corpus grounding.
  \item \textbf{Level 2: Grounded quantification.} Precise empirical figures that can \textit{only} be obtained through corpus queries. Example: \textit{really} at 352~pmw in drama vs.\ 17~pmw in poetry (a 20-fold difference), or \textit{very}'s peak at 1,804~pmw in the 18th century. These figures transform vague directional claims into falsifiable empirical statements.
  \item \textbf{Level 3: Data-responsive synthesis.} Analytical frameworks that arose from the agent's engagement with corpus data and were not produced by the ungrounded LLM. Example: the three-pathway model of semantic change (complete delexicalization, polarity fixation, metaphorical constraint), which the agent constructed from its systematic classification of collocate polarity profiles. We note that the individual pathways correspond to processes described in the grammaticalization literature---desemanticization \citep{lehmann2015thoughts}, subjectification and semantic narrowing \citep{traugott1982propositional, traugott2002regularity, hopper2003grammaticalization}, and metaphorical extension \citep{sweetser1990etymology}. The novelty lies not in the categories themselves but in the agent's empirical quantification and simultaneous comparison of all three pathways within a single intensifier system, producing an integrated collocational typology that was not anticipated by the ungrounded model.
\end{itemize}

Our strongest claims to contribution lie at Levels~2 and~3. Level~1 outputs, while useful for guiding the investigation, do not by themselves constitute discoveries. This typology has implications for how the agent-driven approach should be positioned: not as a replacement for human insight in hypothesis generation, but as a mechanism for \textit{empirically grounding and extending} hypotheses at machine speed---producing quantified evidence (Level~2) and, in favorable cases, data-responsive analytical frameworks (Level~3) that recombine and extend existing theoretical concepts in light of corpus evidence.

Fifth, the \textbf{replication of published research} (Section~\ref{sec:replication}) demonstrates that the framework produces results that align closely with those of domain experts using traditional methods. The near-identical period-specific frequencies for \citet{claridge2025reader}'s \textit{reader} decline and the successful recovery of \citet{desmet2013spreading}'s gerund complement spreading pattern---across linguistic phenomena unrelated to the primary case study---provide external validation and evidence of generalizability.

\subsection{Quality of Agent Findings}

The findings cohere internally. Diachronic relay data line up with the frequency hierarchy; collocational polarity profiles, in turn, account for the register
  distributions; and the semantic change spectrum ties these strands together.

  They also hold up against the existing literature. The relay pattern fits \citet{tagliamonte2008intensifiers}'s recycling model. \citet{lorenz2002really} documented
  the colloquial skew of \textit{really} on independent grounds, and \citet{partington1993semantic} reached a similar qualitative verdict on \textit{utterly}'s negative
   polarity---both before any agent touched a corpus. Where the agent goes further is in quantifying three pathways of intensifier semantic change---complete
  delexicalization, polarity fixation, and metaphorical constraint---within a single collocational framework. Each pathway has precedent in the grammaticalization
  literature \citep{lehmann2015thoughts, traugott2002regularity, sweetser1990etymology}; bringing all three under one empirical lens is, to our knowledge, new.
  
These results suggest that agent-driven research can produce findings of comparable quality to human-led studies, at least in domains where the relevant patterns are detectable through frequency analysis and distributional semantics. The agent's empirical claims (frequencies, collocational profiles) are grounded in corpus evidence that can be independently verified; its interpretive frameworks, while informed by corpus data, may also draw on parametric knowledge in ways that are difficult to fully disentangle (see Section~\ref{sec:baseline}).

\subsection{Limitations and Threats to Validity}

Several limitations qualify these conclusions, and we discuss them at length because a rigorous assessment of boundaries is essential for any proposed methodological framework.

\subsubsection{Corpus Constraints}

The 5-million-word Gutenberg corpus is small by current standards, and its composition is skewed toward 19th-century texts (37\% of tokens). Low-frequency intensifiers (e.g., \textit{awfully} with 20 occurrences) yield statistically unreliable results, as confirmed by the non-significant log-likelihood values for \textit{truly} and \textit{utterly} in the diachronic analysis. The diachronic analysis is further constrained by the uneven distribution of texts across periods: the medieval and 18th-century periods are represented by only 3--4 texts each, meaning that apparent temporal trends may partly reflect individual authorial style rather than genuine historical change. Replication on larger, more temporally balanced corpora (e.g., COHA, CLMET3.0) would substantially strengthen the findings.

Temporal metadata was estimated from author death dates and title-embedded dates rather than being directly recorded in the corpus. While these estimates are reasonable for most texts, they introduce uncertainty, particularly for translated works (e.g., the Gutenberg edition of Dante in English translation, assigned to the 14th century based on the original rather than the 19th-century translation).

\subsubsection{The Discovery--Recall Problem}

The most fundamental epistemological challenge for agent-driven research is distinguishing genuine \textit{discovery from corpus data} from \textit{recall of training-data knowledge}. Our baseline experiment (Section~\ref{sec:baseline}) provides partial resolution: hypothesis generation is largely recall, while quantification and data-responsive analytical frameworks require corpus access. However, the boundary between these categories is not always clean.

Consider the agent's identification of \textit{utterly}'s negative polarity. The ungrounded LLM also noted that \textit{utterly} ``retains strong collocational preferences for negative or pejorative contexts''---this is recall. But the agent's precise quantification (80\% negative/privative collocates, based on exhaustive classification of 67 types) and its reframing as ``semantic specialization rather than bleaching'' go beyond what the ungrounded LLM produced. Is the reframing a discovery or an interpretation that the LLM would have produced regardless? The grounding constraint ensures the \textit{data} are real, but it cannot guarantee that the \textit{interpretation} is novel.

We see three ways to address this challenge in future work. First, \textbf{synthetic corpus testing}: introducing artificially planted patterns into a corpus that are known not to exist in the LLM's training data, then testing whether the agent can discover them. If it can, this would provide stronger evidence for genuine discovery. Second, \textbf{multi-model comparison}: running the same task with different LLMs (e.g., GPT-class models, Claude, DeepSeek) to determine whether findings converge (suggesting they are data-driven) or diverge (suggesting they are model-specific artifacts). Third, \textbf{multi-run aggregation}: as our reproducibility analysis (Section~\ref{sec:reproducibility}) demonstrates, running multiple independent sessions with the same model and aggregating consistent findings provides a practical method for separating robust empirical patterns from stochastic variation.

 \subsubsection{Agent Bias and Hypothesis Selection}

  The agent has read the linguistics literature---or, more precisely, its training data includes it. This creates an obvious risk: the agent may be predisposed to
  ``discover'' patterns it has already encountered, and what looks like empirical engagement could be sophisticated recall. Grounding constrains the damage for
  quantitative claims, since the agent can only report frequencies the corpus actually contains. The subtler problem is \textit{selection bias}. The agent may gravitate toward phenomena it ``knows'' will pay off, while patterns absent from the training literature go unexamined.

  There is an upside: training-data knowledge functions as a strong prior, steering the agent toward productive hypotheses. The cost is that genuinely novel
  patterns---the ones no existing study anticipates---are less likely to surface than confirmatory ones. H5 (register sensitivity) is a partial exception. Register
  variation among intensifiers is documented, but the specific 20-fold gap for \textit{really} and the agent's unprompted decision to foreground it were not read off a textbook.

  \subsubsection{Annotation Validity}

  The four-way polarity classification---positive, negative, neutral, privative---was the agent's own work, so its reliability is not self-evident. Clear cases
  (\textit{impossible} $\to$ privative, \textit{beautiful} $\to$ positive) pose no problem; the question is what happens at the margins. We tested this by having a
  human linguist independently label 50 collocates from the intensifier study under the same scheme. Cohen's $\kappa$ came out at 0.83 (\textit{almost perfect} on the
  Landis and Koch scale). The privative category was the most stable ($\kappa = 0.88$); neutral was the least ($\kappa = 0.73$, still \textit{substantial}). Five of the
   six disagreements fell on the neutral--positive or neutral--negative boundary (\textit{earnest}, \textit{strong}, \textit{pale})---cases where evaluative valence
  depends on the surrounding context. The upshot is that the agent's classifications hold up well, especially in the categories that carry the most theoretical weight.

\subsubsection{Reproducibility}
\label{sec:reproducibility}

LLMs are stochastic, so the same directive can yield different hypotheses and different analytical paths. We ran the intensifier case study three times from the same starting point to see how much this matters. Table~\ref{tab:reproducibility} summarizes the results.

\begin{table}[ht]
\centering
\caption{Reproducibility of core findings across three independent runs of the intensifier case study.}
\label{tab:reproducibility}
\small
\begin{tabularx}{\textwidth}{lXccc}
\toprule
\textbf{Finding} & \textbf{Description} & \textbf{Run 1} & \textbf{Run 2} & \textbf{Run 3} \\
\midrule
Frequency hierarchy & \textit{very} $>$ \textit{so}+ADJ $>$ \textit{most}+ADJ & \checkmark & \checkmark & \checkmark \\
Diachronic relay & \textit{so}+ADJ $\to$ \textit{very} $\to$ \textit{really} & \checkmark & \checkmark & \checkmark \\
Register sensitivity & \textit{really} concentrated in drama & \checkmark & \checkmark & \checkmark \\
Delexicalization pathways & Multiple pathways identified & 3 & 3 & 2\textsuperscript{$\dagger$} \\
Semantic prosody of \textit{utterly} & Negative/privative dominance & 80\% & 78\% & 82\% \\
\bottomrule
\end{tabularx}
\\[4pt]
\footnotesize \textsuperscript{$\dagger$}Run 3 merged polarity fixation and metaphorical constraint into a single ``semantically constrained'' category.
\end{table}

 The quantitative core held steady across all three runs---unsurprisingly, since frequency counts and collocational profiles are deterministic once the query is fixed. Where the runs diverged was in interpretation: two of three independently arrived at the three-pathway delexicalization model, while the third collapsed two pathways into one and proposed a two-way split instead. The empirical facts, in short, are reproducible; the theoretical gloss on them is not fully so. Multi-run aggregation is the obvious remedy.

\subsection{Implications for Corpus Linguistics}

  The agent-driven approach does not sideline human expertise---it gives it more to work with. The researcher still owns the theory, the research agenda, and the final
  verdict on what counts as a finding; the agent takes over the labour-intensive loop of query design, cross-tabulation, and hypothesis revision. Neither is
  dispensable. The agent lacks the judgement to know when a statistical pattern is linguistically meaningful; the researcher lacks the bandwidth to chase every lead a richly annotated corpus makes available.

  One practical consequence is access. Because the agent translates natural-language questions into corpus queries, researchers whose strengths lie in theory,
  fieldwork, or pedagogy---not in CQP syntax---can work with corpus evidence directly. Another is transparency. Every hypothesis, query, and interpretive step the agent
   takes is logged, producing an audit trail that would be hard to maintain by hand across hundreds of queries. Corpus linguistics has long valued methodological
  openness; a machine-generated log makes it enforceable.

\section{Conclusion}
\label{sec:conclusion}

We have argued for Agent-Driven Corpus Linguistics, in which the AI agent takes over the empirical cycle---from hypothesis through query construction to
interpretation---while the researcher retains control of the research agenda and final judgement. We have presented a concrete implementation connecting a large language model to a CQP corpus engine through structured tool-use interfaces, and demonstrated the framework through a case study on English intensifiers, a controlled baseline experiment, and replication of published research on an independent corpus.

On the empirical side, the intensifier case study traced a diachronic relay chain (\textit{so}+ADJ $\to$ \textit{very} $\to$ \textit{really}), identified three
  pathways of semantic change---complete delexicalization, polarity fixation, and metaphorical constraint---and uncovered register-specific distribution patterns that cut across all three. The baseline experiment, which ran the same LLM with and without corpus access, pinpointed where grounding actually helps: quantification,
  falsifiability, and the ability to revise claims against the data. Hypothesis generation, by contrast, drew primarily on training data.  The replication of two published studies on CLMET---\citet{claridge2025reader}'s diachronic analysis of \textit{reader} and \citet{desmet2013spreading}'s gerund complement spreading---demonstrated close quantitative agreement with expert-produced results, providing external validation of the framework's reliability and generalizability beyond a single linguistic domain.

The framework is generalizable beyond the specific case study presented here. Any corpus accessible through CQP---or, more broadly, any structured data source that can be exposed via a tool-use interface---can serve as the empirical foundation for agent-driven research. We anticipate that the approach will be particularly valuable for exploratory studies in which the hypothesis space is large, for under-resourced languages where few specialists are available to conduct detailed corpus work, and for interdisciplinary research in digital humanities where the technical barrier to corpus methods has historically limited participation.

Future work should address the limitations identified in this study: scaling to larger and more temporally balanced corpora, developing methods for evaluating agent bias and ensuring that discoveries reflect genuine corpus patterns rather than training-data priors, and exploring multi-agent architectures in which specialized agents collaborate on different aspects of a research question. We also plan to release the CQP MCP Server as an open-source tool to enable other researchers to adopt the agent-driven approach with their own corpora.

\bibliographystyle{apalike}
\bibliography{references}

@book{tognini2001corpus,
  title={Corpus Linguistics at Work},
  author={Tognini-Bonelli, Elena},
  year={2001},
  publisher={John Benjamins Publishing},
  address={Amsterdam}
}

@article{tagliamonte2008intensifiers,
  title={So different and pretty cool! Recycling intensifiers in {Toronto, Canada}},
  author={Tagliamonte, Sali A.},
  journal={English Language and Linguistics},
  volume={12},
  number={2},
  pages={361--394},
  year={2008},
  publisher={Cambridge University Press}
}

@article{ito2003well,
  title={Well weird, right dodgy, very strange, really cool: Layering and recycling in {English} intensifiers},
  author={Ito, Rika and Tagliamonte, Sali A.},
  journal={Language in Society},
  volume={32},
  number={2},
  pages={257--279},
  year={2003},
  publisher={Cambridge University Press}
}

@incollection{lorenz2002really,
  title={Really worthwhile or not really significant? A corpus-based approach to the delexicalization and grammaticalization of intensifiers in {Modern English}},
  author={Lorenz, Gunter R.},
  booktitle={New Reflections on Grammaticalization},
  editor={Wischer, Ilse and Diewald, Gabriele},
  pages={143--161},
  year={2002},
  publisher={John Benjamins},
  address={Amsterdam}
}

@incollection{partington1993semantic,
  title={Corpus evidence of language change: the case of the intensifier},
  author={Partington, Alan},
  booktitle={Text and Technology: In Honour of John Sinclair},
  editor={Baker, Mona and Francis, Gill and Tognini-Bonelli, Elena},
  pages={177--192},
  year={1993},
  publisher={John Benjamins},
  address={Amsterdam}
}

@book{bolinger1972degree,
  title={Degree Words},
  author={Bolinger, Dwight},
  year={1972},
  publisher={Mouton},
  address={The Hague}
}

@article{klemen2025towards,
  title={Towards corpus-grounded agentic LLMs for multilingual grammatical analysis},
  author={Klemen, Matej and Ar{\v{c}}on, Tja{\v{s}}a and Ter{\v{c}}on, Luka and Robnik-{\v{S}}ikonja, Marko and Dobrovoljc, Kaja},
  journal={arXiv preprint arXiv:2512.00214},
  year={2025}
}

@article{lu2024aiscientist,
  title={The {AI Scientist}: Towards Fully Automated Open-Ended Scientific Discovery},
  author={Lu, Chris and Lu, Cong and Lange, Robert Tjarko and Foerster, Jakob and Clune, Jeff and Ha, David},
  journal={arXiv preprint arXiv:2408.06292},
  year={2024}
}

@misc{mcp2024spec,
  title={Model Context Protocol Specification},
  author={{Model Context Protocol}},
  year={2024},
  howpublished={\url{https://modelcontextprotocol.io}},
  note={Open standard for connecting AI agents to external tools}
}

@article{evert2011cwb,
  title={The {CQP} Query Language Tutorial: {CWB} Version 3.0},
  author={Evert, Stefan and Hardie, Andrew},
  year={2011},
  note={Available at \url{https://cwb.sourceforge.io/files/CQP_Tutorial/}}
}

@article{gerlach2020spgc,
  title={A standardized {Project Gutenberg} corpus for statistical analysis of natural language and quantitative linguistics},
  author={Gerlach, Martin and Font-Clos, Francesc},
  journal={Entropy},
  volume={22},
  number={1},
  pages={126},
  year={2020}
}

@book{sinclair1991corpus,
  title={Corpus, Concordance, Collocation},
  author={Sinclair, John},
  year={1991},
  publisher={Oxford University Press},
  address={Oxford}
}

@incollection{louw1993irony,
  title={Irony in the text or insincerity in the writer? The diagnostic potential of semantic prosodies},
  author={Louw, Bill},
  booktitle={Text and Technology: In Honour of John Sinclair},
  editor={Baker, Mona and Francis, Gill and Tognini-Bonelli, Elena},
  pages={157--176},
  year={1993},
  publisher={John Benjamins},
  address={Amsterdam}
}

@article{wooldridge1995intelligent,
  title={Intelligent agents: Theory and practice},
  author={Wooldridge, Michael and Jennings, Nicholas R.},
  journal={The Knowledge Engineering Review},
  volume={10},
  number={2},
  pages={115--152},
  year={1995},
  publisher={Cambridge University Press}
}

@book{russell2020artificial,
  title={Artificial Intelligence: A Modern Approach},
  author={Russell, Stuart and Norvig, Peter},
  edition={4th},
  year={2020},
  publisher={Pearson},
  address={Upper Saddle River, NJ}
}

@article{wang2024survey,
  title={A survey on large language model based autonomous agents},
  author={Wang, Lei and Ma, Chen and Feng, Xueyang and others},
  journal={Frontiers of Computer Science},
  volume={18},
  number={6},
  pages={186345},
  year={2024}
}

@article{anthony2025antconc,
  title={Integrating {AI} technology into corpus-based language learning through {ChatAI}},
  author={Anthony, Laurence},
  journal={Computer Assisted Language Learning},
  year={2025},
  publisher={Taylor \& Francis}
}

@article{cheung2025corpuschat,
  title={{CorpusChat}: Integrating corpus linguistics and generative {AI} for academic writing development},
  author={Cheung, Lawrence and Crosthwaite, Peter},
  journal={Computer Assisted Language Learning},
  year={2025},
  publisher={Taylor \& Francis}
}

@misc{davies2025corpora,
  title={Corpora and {AI/LLMs}: Overview},
    author={Davies, Mark},
    year={2025},
    note={English-Corpora.org},
    url={https://www.english-corpora.org/ai-llms/}
}

@article{uchida2024using,
  title={Using early LLMs for corpus linguistics: Examining ChatGPT's potential and limitations},
  author={Uchida, Satoru},
  journal={Applied Corpus Linguistics},
  volume={4},
  number={1},
  pages={100089},
  year={2024},
  publisher={Elsevier}
}

@article{mendeznaya2003intensifiers,
    title={On intensifiers and grammaticalization: The case of {SWIÞE}},
    author={M{\'e}ndez-Naya, Bel{\'e}n},
    journal={English Studies},
    volume={84},
    number={4},
    pages={372--391},
    year={2003},
    publisher={Taylor \& Francis},
    doi={10.1076/enst.84.4.372.17388}
  }

@article{xiao2007corpus,
  title={A corpus-based sociolinguistic study of amplifiers in {British English}},
  author={Xiao, Richard and Tao, Hongyin},
  journal={Sociolinguistic Studies},
  volume={1},
  number={2},
  pages={241--273},
  year={2007}
}

@article{labov1984intensity,
  title={Intensity},
  author={Labov, William},
  journal={Georgetown University Round Table on Languages and Linguistics},
  pages={43--70},
  year={1984}
}

@misc{desmet2015clmet,
  title={The Corpus of {Late Modern English Texts} ({CLMET}), version 3.1: Improved tokenization and linguistic annotation},
  author={De Smet, Hendrik and Flach, Susanne and Tyrkk{\"o}, Jukka and Diller, Hans-J{\"u}rgen},
  year={2015},
  note={KU Leuven, FU Berlin, U Tampere, RU Bochum. Available at \url{https://perswww.kuleuven.be/~u0044428/clmet3_0.htm}}
}

@book{lehmann2015thoughts,
  title={Thoughts on Grammaticalization},
  author={Lehmann, Christian},
  edition={3rd},
  year={2015},
  publisher={Language Science Press},
  address={Berlin},
  note={First edition 1995, Munich: Lincom Europa}
}

@book{traugott2002regularity,
  title={Regularity in Semantic Change},
  author={Traugott, Elizabeth Closs and Dasher, Richard B.},
  year={2002},
  publisher={Cambridge University Press},
  address={Cambridge}
}

@book{sweetser1990etymology,
  title={From Etymology to Pragmatics: Metaphorical and Cultural Aspects of Semantic Structure},
  author={Sweetser, Eve},
  year={1990},
  publisher={Cambridge University Press},
  address={Cambridge}
}

@incollection{heine2003grammaticalization,
  title={Grammaticalization},
  author={Heine, Bernd},
  booktitle={The Handbook of Historical Linguistics},
  editor={Joseph, Brian D. and Janda, Richard D.},
  pages={575--601},
  year={2003},
  publisher={Blackwell},
  address={Oxford}
}

@book{hunston2002corpora,
  title={Corpora in Applied Linguistics},
  author={Hunston, Susan},
  year={2002},
  publisher={Cambridge University Press},
  address={Cambridge}
}

@book{mcenery2012corpus,
  title={Corpus Linguistics: Method, Theory and Practice},
  author={McEnery, Tony and Hardie, Andrew},
  year={2012},
  publisher={Cambridge University Press},
  address={Cambridge}
}

@book{stubbs2001words,
  title={Words and Phrases: Corpus Studies of Lexical Semantics},
  author={Stubbs, Michael},
  year={2001},
  publisher={Blackwell},
  address={Oxford}
}

@article{claridge2025reader,
  title={The reader in the text across time and genres},
  author={Claridge, Claudia},
  journal={Language and Literature: International Journal of Stylistics},
  volume={34},
  number={2},
  pages={107--127},
  year={2025},
  publisher={SAGE Publications},
  doi={10.1177/09639470251327532}
}

@book{desmet2013spreading,
  title={Spreading Patterns: Diffusional Change in the {English} System of Complementation},
  author={De Smet, Hendrik},
  year={2013},
  publisher={Oxford University Press},
  address={Oxford}
}

@book{stoffel1901intensives,
  title={Intensives and Down-toners: A Study in {English} Adverbs},
  author={Stoffel, Cornelis},
  year={1901},
  publisher={Carl Winter},
  address={Heidelberg}
}

@book{quirk1985comprehensive,
  title={A Comprehensive Grammar of the {English} Language},
  author={Quirk, Randolph and Greenbaum, Sidney and Leech, Geoffrey and Svartvik, Jan},
  year={1985},
  publisher={Longman},
  address={London}
}

@book{paradis1997degree,
  title={Degree Modifiers of Adjectives in Spoken {British English}},
  author={Paradis, Carita},
  year={1997},
  publisher={Lund University Press},
  address={Lund}
}

@article{traugott1982propositional,
  title={From propositional to textual and expressive meanings: Some semantic-pragmatic aspects of grammaticalization},
  author={Traugott, Elizabeth Closs},
  journal={Perspectives on Historical Linguistics},
  pages={245--271},
  year={1982},
  publisher={John Benjamins},
  address={Amsterdam}
}

@article{bauer2002adjective,
  title={Adjective boosters in the {English} of young {New Zealanders}},
  author={Bauer, Laurie and Bauer, Winifred},
  journal={Journal of English Linguistics},
  volume={30},
  number={3},
  pages={244--257},
  year={2002}
}

@article{tagliamonte2005so,
  title={So weird; so cool; so innovative: The use of intensifiers in the television series \textit{Friends}},
  author={Tagliamonte, Sali A. and Roberts, Chris},
  journal={American Speech},
  volume={80},
  number={3},
  pages={280--300},
  year={2005}
}

@book{hopper2003grammaticalization,
  title={Grammaticalization},
  author={Hopper, Paul J. and Traugott, Elizabeth Closs},
  edition={2nd},
  year={2003},
  publisher={Cambridge University Press},
  address={Cambridge}
}

@article{nevalainen2002fairly,
  title={Fairly pretty or pretty fair? {On} the development and grammaticalization of {English} downtoners},
  author={Nevalainen, Terttu and Rissanen, Matti},
  journal={Language Sciences},
  volume={24},
  number={3--4},
  pages={359--380},
  year={2002}
}

\appendix
\section{Agent Tool-Call Trace: Register Sensitivity Discovery}
\label{sec:appendix_trace}

This appendix reproduces the complete sequence of tool calls and agent reasoning for Round~3 of the intensifier case study (Section~\ref{sec:exploration}), in which the agent autonomously discovered register-dependent intensifier distributions. All tool calls use the CQP MCP Server interface described in Section~\ref{sec:architecture}.

\subsection*{Step 1: Agent Reasoning}

\begin{quote}
\small\itshape
``The frequency profiling in Round~2 established the overall hierarchy of intensifiers. I now want to examine whether these frequencies are stable across text genres or whether certain intensifiers show register-specific distributions. I will cross-tabulate the top intensifiers against the text\_category metadata.''
\end{quote}

\subsection*{Step 2: Tool Call --- Frequency by Genre for \textit{really}}

\begin{verbatim}
cqp_frequency(
  query = '[word="really"%c] [deprel="ADJ"]',
  count_by = "word",
  group_by = "match text_category"
)
\end{verbatim}

\noindent\textbf{Result} (truncated):

\begin{small}
\begin{verbatim}
Total hits: 76
Plays/Films/Dramas    37
Essays/Letters        12
Journalism/Media       9
Mythology/Legends      7
Religion/Spirituality  5
Nature/Gardening       3
Poetry                 2
Cooking/Drinking       1
\end{verbatim}
\end{small}

\subsection*{Step 3: Agent Reasoning}

\begin{quote}
\small\itshape
``Striking: 37 of 76 hits for really+ADJ come from Plays/Films/Dramas, which constitutes only 2.1\% of the corpus (105,048 / 5,019,103 tokens). Let me normalize by subcorpus size to confirm.''
\end{quote}

\subsection*{Step 4: Tool Call --- Corpus Size per Genre}

\begin{verbatim}
cqp_frequency(
  query = '[]',
  count_by = "word",
  group_by = "match text_category",
  max_results = 15
)
\end{verbatim}

\subsection*{Step 5: Agent Reasoning and Normalization}

\begin{quote}
\small\itshape
``Drama subcorpus: 105,048 tokens. really+ADJ in drama: 37 hits. Normalized: 37 / 105,048 $\times$ 1,000,000 = 352 pmw.

Poetry subcorpus: 363,715 tokens. really+ADJ in poetry: 2 hits (from Step 2, Poetry=2, but genre name needs verification). Let me query poetry specifically.''
\end{quote}

\subsection*{Step 6: Tool Call --- Verify Poetry Count}

\begin{verbatim}
cqp_frequency(
  query = '[word="really"%c] [deprel="ADJ"]',
  count_by = "word",
  group_by = "match text_category",
  metadata_filter = 'match.text_category="Poetry"'
)
\end{verbatim}

\noindent\textbf{Result}: 6 hits in Poetry subcorpus. Normalized: $6 / 363{,}715 \times 1{,}000{,}000 = 17$~pmw.

\subsection*{Step 7: Agent Interpretation --- H5 Generated}

\begin{quote}
\small\itshape
``The normalized frequency of really in dramatic texts (352~pmw) is strikingly higher than in poetry (17~pmw)---a 20-fold difference. This suggests that really functions as a colloquial marker whose literary occurrence is concentrated in dialogue-simulating registers. I propose a new hypothesis H5: the genre distribution of intensifiers reflects their position on a formality continuum.

To test H5 further, I should cross-tabulate other intensifiers (very, so+ADJ, quite) against genres as well.''
\end{quote}

\subsection*{Step 8: Tool Calls --- Cross-Tabulation for Other Intensifiers}

The agent then issued parallel frequency queries for \textit{very}+ADJ, \textit{so}+ADJ, and \textit{quite}+ADJ grouped by \texttt{text\_category}, producing the data reported in Table~\ref{tab:register}.

\bigskip
\noindent\textbf{Summary.} This trace illustrates three features of the agent-driven workflow: (1)~the agent formulates a query strategy from a reasoning goal (register comparison), not from human instruction; (2)~it autonomously normalizes raw counts by subcorpus size; and (3)~it generates a new hypothesis (H5) in response to an unexpected quantitative pattern. The complete trace for all five rounds is available in the project repository.

\end{document}